\documentclass[letterpaper,10pt,conference]{ieeeconf}
\IEEEoverridecommandlockouts
\usepackage{amsmath}
\usepackage{amssymb}
\usepackage{graphicx}
\usepackage{booktabs}
\usepackage{multirow}
\usepackage{caption}
\usepackage{subcaption}
\usepackage{float}
\usepackage{tikz}
\usepackage{mdframed}
\begin{document}

\title{\LARGE \bf SNR-Adaptive Unified Diffusion for Multi-Task Medical Image Segmentation}

\author{
    Jiahao Liu$^1$, Hang Wei$^{2*}$, Shuai Wu$^{2*}$ \\
    $^1$School of Telecommunications Engineering, Xidian University \\
    $^2$School of Computer Science and Technology, Xidian University \\
    \texttt{23012100032@stu.xidian.edu.cn, \{weihang, wushuai\}@xidian.edu.cn}
}

\maketitle
\thispagestyle{empty}
\pagestyle{empty}

\begin{abstract}
Clinical cardiac imaging pipelines currently deploy separate models for
each dataset and modality, incurring redundant training costs and
precluding knowledge sharing across anatomically related tasks.
Consolidating semi-supervised learning, unsupervised domain adaptation,
and domain generalisation into one model is therefore a practical
necessity, yet naive joint training exposes a fundamental barrier:
conflicting label semantics between datasets collapse LA Dice from
90.31\% to 83.38\%, while gradient imbalance across tasks of
unequal complexity suppresses the weaker tasks throughout training.
We present UniT-Diff, a unified diffusion segmentation framework that
resolves these conflicts through three targeted mechanisms.
An 11-channel task-specific output space physically partitions label
categories, eliminating cross-task gradient sign reversal by
construction.
SNR-Adaptive Task Conditioning (SATC) scales the task token by the
log signal-to-noise ratio of the current diffusion timestep,
suppressing domain-specific bias during coarse denoising and restoring
full task guidance as the signal clears.
Task-Type-Aware Conditional Dropout (TTACD) permanently removes the
task token for domain-generalisation inputs, routing them through a
shared neutral pathway that draws on cross-dataset cardiac anatomy
rather than source-vendor statistics.
Under a single parameter set, UniT-Diff surpasses independently trained
task-specific baselines on all three benchmarks simultaneously:
+0.87\% on LA, +1.77\% on MMWHS, and +0.88\% on MNMS.

\end{abstract}

\section{Introduction}
\label{sec:introduction}

Medical image segmentation underpins quantitative analysis, surgical
planning, and longitudinal monitoring across modern clinical
workflows~\cite{isensee2021nnunet}.
In practice, a hospital's imaging pipeline must contend with
MRI acquired from multiple vendors, CT collected under varying
contrast protocols, and multi-centre cohorts whose annotation
conventions differ organ by organ.
The prevailing response is to maintain a separate model for each
dataset and modality---a ``one-model-per-task'' paradigm whose
storage overhead, version management burden, and inability to share
anatomical knowledge across related structures scale poorly as
data diversity grows.
Consolidating semi-supervised learning (SSL), unsupervised domain
adaptation (UDA), and domain generalisation (DG) into a single
deployable parameter set is therefore not merely an academic exercise
but a practical clinical necessity.

\begin{figure}[!t]
\centering
\begin{tikzpicture}
  \node[draw, rounded corners=6pt, line width=0.5pt,
        inner sep=6pt, fill=white] {%
    \begin{minipage}{0.93\columnwidth}
      \centering
      \includegraphics[width=\linewidth]{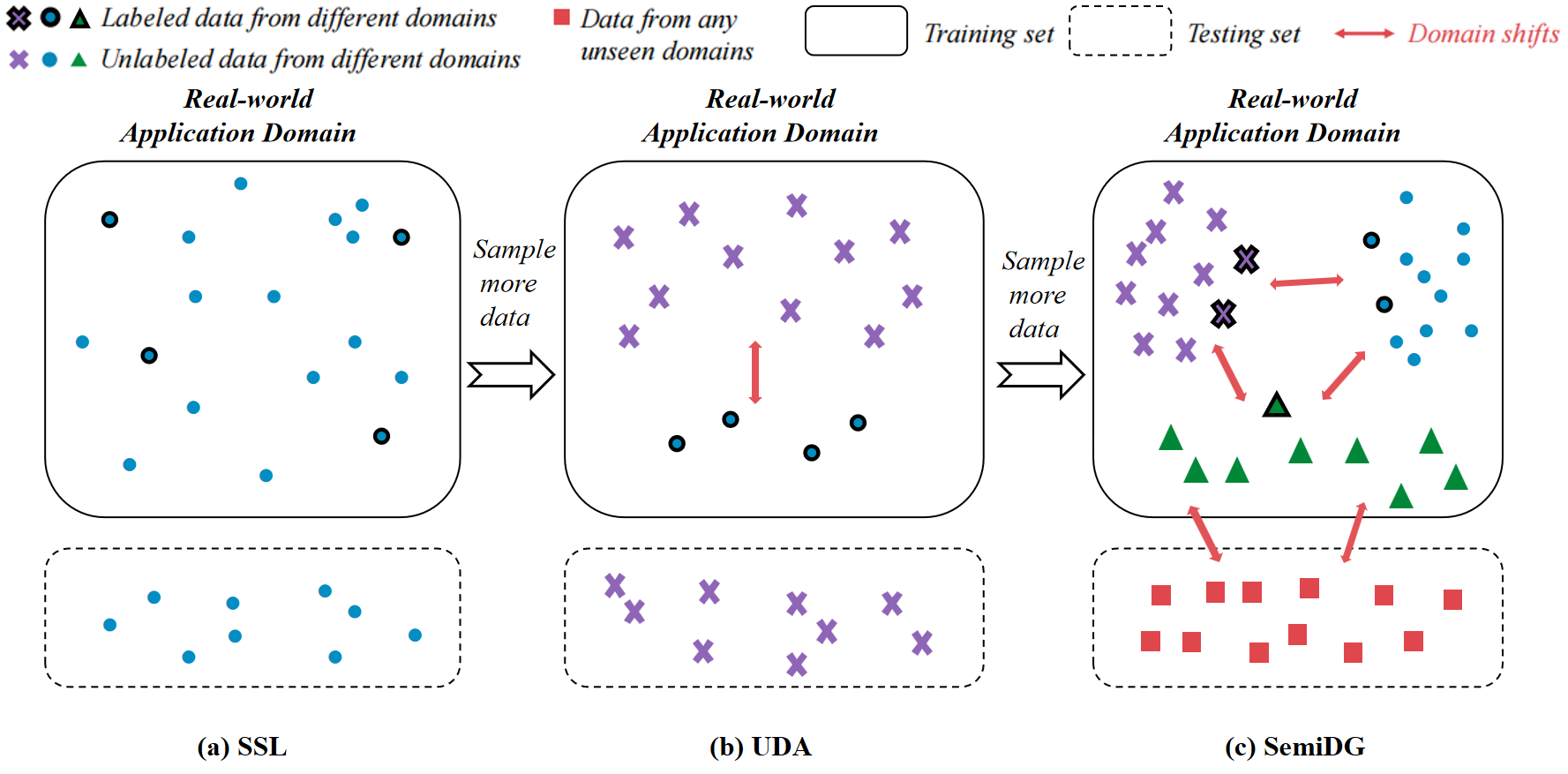}
      \vspace{2pt}
      {\small\textbf{(a)} SSL, UDA, and SemiDG represent increasing
      levels of distributional diversity between training and test data.}
      \par\vspace{5pt}
      \noindent\makebox[\linewidth]{%
        \tikz\draw[dashed, line width=0.4pt, dash pattern=on 4pt off 2pt]
        (0,0) -- (\linewidth,0);}
      \vspace{5pt}
      \includegraphics[width=\linewidth]{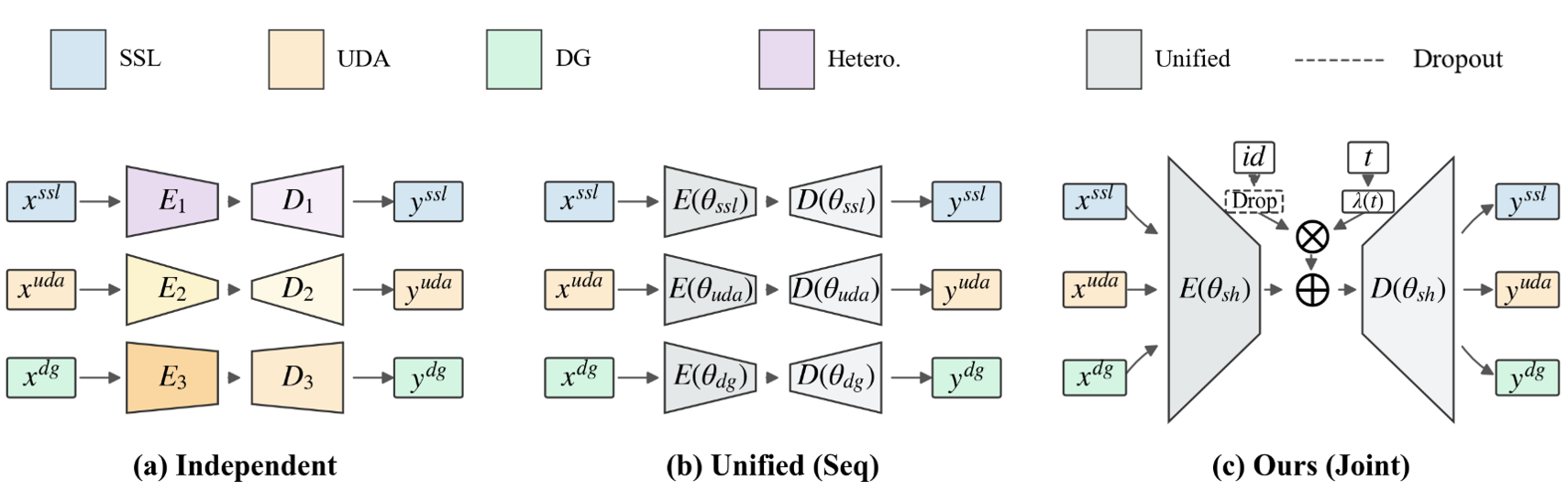}
      \vspace{2pt}
      {\small\textbf{(b)} Training paradigm comparison:
      \textbf{(i)}~Independent single-task models,
      \textbf{(ii)}~GenericSSL baseline,
      \textbf{(iii)}~Proposed UniT-Diff.}
    \end{minipage}%
  };
\end{tikzpicture}
\caption{Overview of the problem setting and proposed paradigm.
\textbf{(a)}~The three tasks span increasing distributional diversity,
making naive joint optimisation unstable.
\textbf{(b)}~UniT-Diff consolidates all three tasks into a single
parameter set, enabling positive knowledge transfer across SSL,
UDA, and SemiDG.}
\label{fig:overview}
\end{figure}

Diffusion probabilistic models have demonstrated strong anatomical
priors in single-task medical segmentation~\cite{ho2020denoising,
wu2022medsegdiff}, and the generic semi-supervised diffusion framework
DiffVNet~\cite{wang2024genericssl} has established state-of-the-art
results on each of the three tasks in isolation.
The central question of this work is whether these single-task
strengths can be preserved---and ideally reinforced---when all three
datasets are trained jointly within DiffVNet's diffusion prior.
A naive answer is to merge all datasets under a shared 9-channel
output head.
The result is immediate and severe: LA Dice collapses from $90.31\%$
to $83.38\%$.
Post-hoc gradient analysis identifies two coupled failure modes.
\emph{Semantic collision} arises because the left ventricle is
foreground in MMWHS but background in LA; any shared output channel
receives opposing gradients from the two tasks simultaneously.
\emph{Gradient imbalance} arises because the five-class MMWHS task
produces systematically larger gradient norms than the two-class LA
task, biasing shared encoder updates toward the more complex dataset.
Fig.~\ref{fig:overview}(a) illustrates how the three tasks span
increasing levels of distribution diversity, which together make
naive joint optimisation unstable.

Beyond these structural conflicts, task conditioning itself introduces
a subtler tension.
A fixed task token injected uniformly throughout the denoising chain
biases the model toward source-domain statistics at every noise level,
even when the latent representation is too coarse to benefit from
task-specific guidance.
Moreover, for domain-generalisation inputs---where the test scanner
is never seen during training---the task token associates samples
with source-vendor statistics rather than generalisable anatomy,
actively degrading out-of-distribution accuracy.
Task conditioning is therefore simultaneously a necessity for
cross-modal alignment and a liability for cross-centre generalisation,
depending on which task is being processed.

Building on DiffVNet~\cite{wang2024genericssl} as the backbone,
this work investigates how to fuse three heterogeneous cardiac
datasets into a single joint training pipeline without negative
transfer.
Fig.~\ref{fig:overview}(b) summarises the resulting framework,
UniT-Diff, relative to prior paradigms.
The contributions are threefold.

\begin{itemize}

\item \textbf{11-channel unified label space for joint optimisation.}
For the first time in diffusion-based medical image segmentation, we
demonstrate that SSL, UDA, and DG can be optimised jointly within a
single parameter set. By assigning each task a non-overlapping output
interval, we physically eliminate cross-task gradient sign reversal,
enabling truly simultaneous training without task-specific network
branches or sequential task switching.

\item \textbf{SNR-Adaptive Task Conditioning (SATC).}
The task embedding is gated by the log signal-to-noise ratio of
the current diffusion timestep via a per-task learnable temperature,
suppressing domain-specific bias during coarse denoising and
restoring full task guidance as the signal-to-noise ratio increases.

\item \textbf{Task-Type-Aware Conditional Dropout (TTACD).}
Token dropout probabilities are assigned according to each task's
learning objective---$20\%$ for SSL (LA) combined with loss
re-weighting to compensate for limited labeled data, $0\%$ for
UDA (MMWHS) to preserve cross-modal alignment, and $100\%$ for DG
(MNMS) to enforce vendor-agnostic inference through the shared
neutral pathway.

\end{itemize}

Evaluated on LA, MMWHS, and MNMS, UniT-Diff surpasses the
independently trained DiffVNet baselines on all three benchmarks
simultaneously under a single parameter set: $+0.87\%$ on LA,
$+1.77\%$ on MMWHS, and $+0.88\%$ on MNMS.

\section{Related Work}

\subsection{Semi-supervised Medical Image Segmentation}
Reducing annotation dependence in medical segmentation has two dominant research lines. 
Consistency-based and pseudo-labeling methods~\cite{yu2019uncertainty,bai2023bcp,wu2021mcnet} regularize unlabeled predictions via perturbation invariance and label propagation, achieving strong single-task performance under low-label regimes. 
However, all these methods rely on a fixed, homogeneous label space, and cannot resolve annotation conflicts across heterogeneous multi-dataset settings.

Diffusion models have recently emerged as a powerful semi-supervised segmentation framework. 
Our backbone DiffVNet~\cite{wang2024genericssl} frames segmentation as iterative mask denoising, achieving state-of-the-art results via its strong anatomical prior. 
Yet it assumes a unified label space throughout training, which becomes a critical barrier when combining datasets with conflicting class definitions.

\subsection{Domain Adaptation and Generalisation}
Cross-domain cardiac segmentation follows two complementary paradigms. 
Unsupervised domain adaptation (UDA) aligns labeled source and unlabeled target modalities, with recent diffusion-based methods~\cite{yang2024addressing,Gao2026retri} leveraging DDPM priors for cross-modality alignment. 
Domain generalisation (DG) learns vendor-agnostic representations without access to target data at training time~\cite{liu2021dgnet,liu2022vmfnet}.

A shared limitation across both families is that they are designed for a single organ or modality pair. 
No existing mechanism can selectively enable or suppress domain-specific signals based on the task objective, and their training pipelines cannot accommodate additional tasks with conflicting label conventions.

\subsection{Multi-task Learning and Gradient Conflict}
Multi-task learning (MTL) trains a shared encoder across multiple objectives to improve generalisation, but suffers from conflicting gradients when tasks have opposing optimisation targets. 
Gradient surgery methods such as PCGrad~\cite{yu2020projecting} and GradNorm~\cite{chen2018gradnorm} resolve this via gradient projection or adaptive loss reweighting, and are widely used in multi-organ medical segmentation.

A core, unchallenged assumption of existing MTL methods is that explicit task identity signals are uniformly beneficial. 
Our work directly challenges this: we empirically show that task conditioning aids cross-modal alignment in UDA, but degrades out-of-distribution generalisation in DG by encoding source-domain statistics absent at test time. 
Neither gradient surgery nor uniform task conditioning addresses this fundamental asymmetry, which our framework resolves via task-aware conditioning policies.

\section{Methodology}
\label{sec:method}

\subsection{Problem Formulation}

Let $\mathcal{D}=\{(\mathcal{D}_k,\mathcal{C}_k)\}_{k=1}^{K}$ with
$K{=}3$ denote three heterogeneous cardiac benchmarks: the Left Atrium
dataset (LA, semi-supervised learning), the Multi-Modality Whole Heart
Segmentation dataset (MMWHS, unsupervised domain adaptation), and the
Multi-Centre Cardiac dataset (MNMS, domain generalisation).
The class counts $|\mathcal{C}_k|$ differ across tasks:
$|\mathcal{C}_1|{=}2$ (LA), $|\mathcal{C}_2|{=}5$ (MMWHS),
$|\mathcal{C}_3|{=}4$ (MNMS), so a naive shared head of size
$\sum_k|\mathcal{C}_k|{-}(K{-}1){=}9$ would conflate background
classes across datasets.Each $\mathcal{D}_k$ contains a small labeled subset $\mathcal{D}_k^L$
and a substantially larger unlabeled subset $\mathcal{D}_k^U$.
Rather than instantiating separate parameters $\theta_k$ per task,
a single model $f_\theta$ is trained to maximise
\begin{equation}
\max_{\theta}\;\sum_{k=1}^{K}
\mathbb{E}_{(\mathbf{x},\mathbf{y})\sim\mathcal{D}_k^L}
\bigl[\mathrm{Dice}\!\left(
  f_\theta(\mathbf{x})[\,s_k{:}e_k],\;\mathbf{y}
\right)\bigr],
\label{eq:objective}
\end{equation}
where $[s_k,e_k)$ denotes the task-specific output channel interval.
The reverse diffusion process $p_\theta(x_{t-1}|x_t,\mathbf{x})$
reconstructs the segmentation mask conditioned on the input volume
$\mathbf{x}$, following the formulation of
DiffVNet~\cite{wang2024genericssl}.
Fig.~\ref{fig:overview} contrasts our joint training strategy
against single-task and sequential baselines.

\begin{figure}[t]
\centering
\includegraphics[width=\columnwidth]{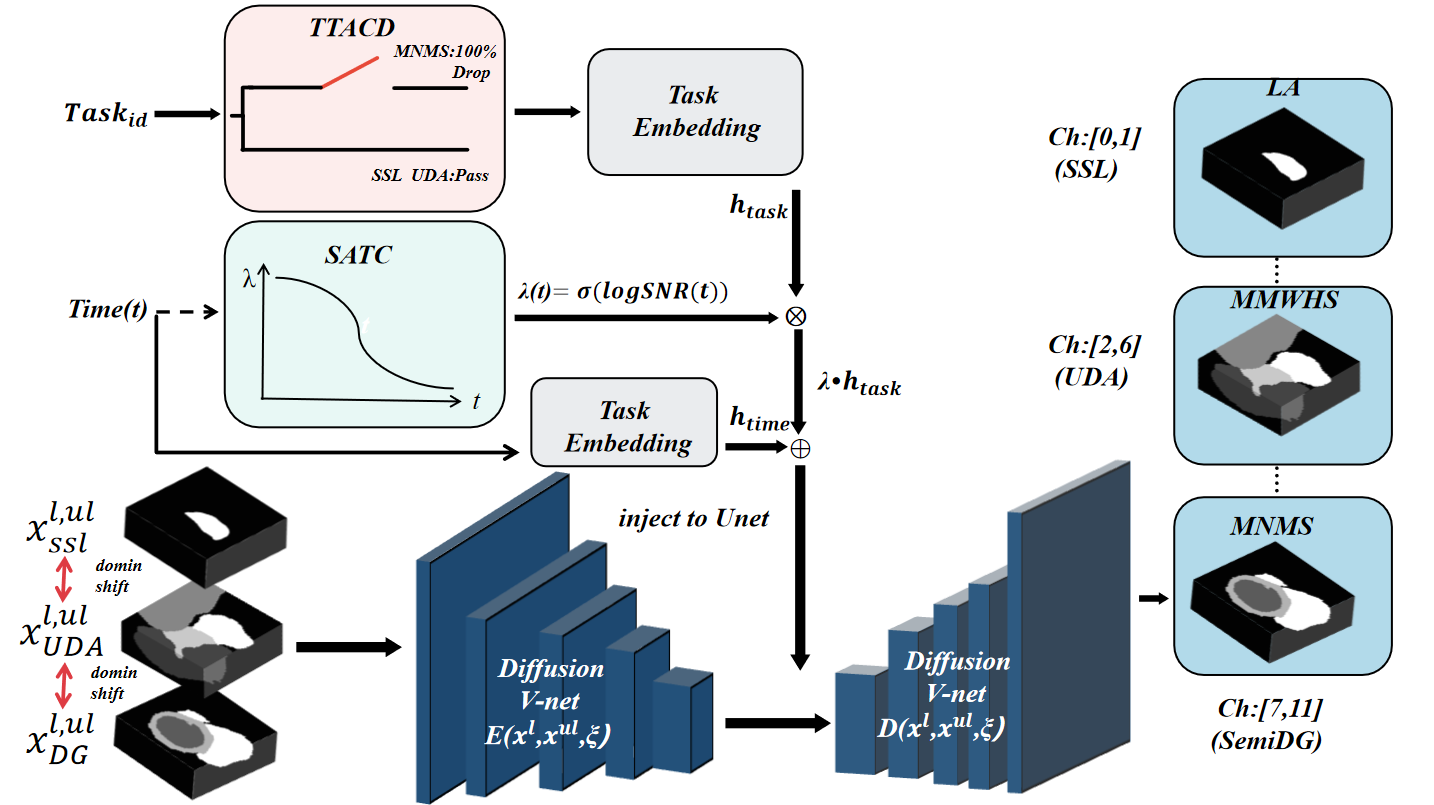}
\caption{Architecture of the proposed UniT-Diff framework.
Three heterogeneous cardiac datasets feed a shared DiffVNet backbone
whose 11-channel output is partitioned into non-overlapping task
intervals to eliminate semantic collision.
SNR-Adaptive Task Conditioning (SATC) injects a learnable task token
scaled by a per-task temperature and the instantaneous log-SNR,
providing strong guidance at low noise levels while suppressing
domain-specific bias during coarse denoising.
Task-Type-Aware Conditional Dropout (TTACD) permanently removes the
task token for the domain-generalisation stream (MNMS), routing those
samples through the shared neutral pathway at inference.
}
\label{fig:framework}
\end{figure}

\subsection{Unified 11-Channel Label Space}
\label{subsec:label_space}

The overall architecture is illustrated in Fig.~\ref{fig:framework}.
A fundamental obstacle in joint training over heterogeneous datasets
is \emph{semantic collision}: the same anatomical region carries
contradictory labels across tasks.
The left ventricle, annotated as foreground in MMWHS, belongs to the
background class in LA.
Any shared output channel therefore receives opposing gradients
$\nabla\mathcal{L}_{\text{LA}}$ and $\nabla\mathcal{L}_{\text{MMWHS}}$
simultaneously, causing destructive interference that degrades both tasks.
A 9-channel design that merges the background class across all tasks
confirms this effect empirically: LA Dice drops from $90.31\%$ to
$83.38\%$, a degradation of $6.93$ percentage points.

To eliminate this interference by construction, the output space is
expanded to 11 non-overlapping channels with exclusive task intervals:
$[0,2)$ for LA, $[2,7)$ for MMWHS, and $[7,11)$ for MNMS.
For a voxel labelled $c$ at position $(h,w,d)$ in task~$k$, the
one-hot encoding populates only the interval $[s_k,e_k)$:
\begin{equation}
y_{\mathrm{oh}}[\,b,\,\tau_k(c),\,h,w,d\,] = 1
\;\;\Leftrightarrow\;\;
\mathrm{label}[\,b,h,w,d\,] = c,
\label{eq:onehot}
\end{equation}
and the diffusion prior is scaled to
$x_{\mathrm{start}}=y_{\mathrm{oh}}\times 2-1\in\{-1,+1\}^{11}$.
Because each task's softmax operates exclusively on its own channels,
gradient sign reversal is structurally prevented.
One consequence of Eq.~\eqref{eq:onehot} is that channels outside
$[s_k,e_k)$ are padded with $-1$ in $x_{\mathrm{start}}$,
encoding them as ``background'' rather than ``unlabelled''---a prior
mismatch whose effect on generative capacity is discussed in
Sect.~\ref{sec:discussion}.

Two implementation details are necessary to realise the benefit of this
design in practice.
First, three independent loss instances must be maintained per task;
sharing a single loss object across batches of varying channel
dimensionality corrupts internally cached class-weight statistics,
producing erratic gradient magnitudes that destabilise convergence.
Second, each task retains an independent \emph{Difficulty}
monitor~\cite{wang2024genericssl} that tracks per-class Dice progress
and dynamically adjusts the supervised loss weight, preserving the
adaptive difficulty-aware training of the single-task baseline within
the unified framework.

\subsection{SNR-Adaptive Task Conditioning}
\label{subsec:satc}

Channel decoupling resolves label-space conflicts but leaves the shared
encoder without any cue as to which input modality it is processing.
Adding a fixed task token to the timestep embedding is the natural
remedy, yet uniform injection across all noise levels is theoretically
unsound.
At high noise levels the latent state retains only coarse, low-frequency
anatomy; a strong task signal at this stage biases the denoising
trajectory toward training-domain statistics before any task-relevant
fine structure is recoverable.
We quantify the noise level via
\begin{equation}
\mathrm{SNR}(t) = \frac{\bar{\alpha}_t}{1-\bar{\alpha}_t},
\label{eq:snr}
\end{equation}
where $\bar{\alpha}_t$ is the cumulative noise schedule~\cite{ho2020denoising}.
The proposed \emph{SNR-Adaptive Task Conditioning} (SATC) gates the task
embedding by the log-SNR:
\begin{equation}
h_{\mathrm{cond}} = h_{\mathrm{time}} +
\underbrace{\sigma\!\left(\tau_k\cdot\log\mathrm{SNR}(t)\right)}_{\lambda_k(t)}
\cdot h_{\mathrm{task}},
\label{eq:satc}
\end{equation}
where $h_{\mathrm{time}}$ is the sinusoidal timestep embedding,
$h_{\mathrm{task}}\in\mathbb{R}^d$ is a learnable task token, and
$\tau_k$ is a \emph{per-task learnable temperature}.
The gate satisfies $\lambda_k(t)\!\to\!0$ as $t\!\to\!T$
(shared anatomy dominates under high noise) and
$\lambda_k(t)\!\to\!1$ as $t\!\to\!0$
(task-specific detail is fully injected near the clean signal).

The per-task parameterisation of $\tau_k$ is central to this design.
MMWHS requires aggressive cross-modality alignment---CT and MRI
intensity distributions differ substantially---and converges to a
steeper gate.
LA, operating entirely within single-centre MRI, benefits from a
shallower slope that avoids over-specialising the shared encoder.
A shared scalar temperature would allow the larger gradient of the
five-class MMWHS task to dictate the conditioning profile of the
two-class LA task, a coupling that consistently harms LA performance
in ablation.
This formulation draws on classifier-free guidance~\cite{ho2022classifier},
where conditioning strength is itself a trainable quantity, and extends
it to the multi-task regime by making that strength task-dependent and
noise-level-aware simultaneously.

\subsection{Task-Type-Aware Conditional Dropout}
\label{subsec:ttacd}

SATC regulates the temporal conditioning profile, but a second,
qualitatively distinct failure mode persists for domain generalisation.
At MNMS inference, any task embedding associates inputs with the source
vendors (B, C, D) seen during training; since the target vendor (A) is
never observed, this vendor-specific association provides no useful
inductive bias and actively degrades out-of-distribution accuracy.
The core tension is that task conditioning is simultaneously necessary
for cross-modal alignment and harmful for cross-centre generalisation,
depending on which task is being processed.

\emph{Task-Type-Aware Conditional Dropout} (TTACD) resolves this
tension by assigning each task a deterministic token-dropout
probability grounded in its learning objective.
Unlike the uniform dropout of classifier-free
guidance~\cite{ho2022classifier}, which treats all conditions
symmetrically, TTACD differentiates by task type:

\begin{table}[h]
\small
\centering
\setlength{\tabcolsep}{5pt}
\caption{Task-Type-Aware Conditional Dropout (TTACD) probabilities.}
\label{tab:dropout}
\begin{tabular}{p{1.1cm}p{0.8cm}p{0.6cm}p{4.2cm}}
\toprule
Dataset & Type & $p_{\mathrm{drop}}$ & Rationale \\
\midrule
LA    & SSL & 20\% & Reinforces neutral path; mild regulariser
                    for the 8-scan labeled set \\[2pt]
MMWHS & UDA &  0\% & Full conditioning for CT$\leftrightarrow$MRI
                    alignment \\[2pt]
MNMS  & DG  & 100\% & Enforces vendor-agnostic inference at test time \\
\bottomrule
\end{tabular}
\end{table}

For MMWHS, any suppression of the task token eliminates the
cross-modality alignment signal central to UDA performance.
For MNMS, permanent suppression---applied identically during training
and inference---forces segmentation through the \emph{neutral pathway},
the shared cardiac representation accumulated from both LA and MMWHS,
without access to source-vendor statistics.
The $20\%$ rate for LA serves a dual purpose: it periodically
reinforces the neutral pathway with single-centre MRI samples,
improving the quality of that path for MNMS inference, and acts as a
mild regulariser on the limited labeled set.

\paragraph{Task loss re-weighting.}
A complementary challenge arises from gradient imbalance across tasks.
LA provides only 8 labeled volumes against MMWHS's 20; the effective
gradient contribution of LA---further reduced by 20\% dropout
episodes---risks being overwhelmed by the more data-rich MMWHS.
A per-task scalar weight $w_k$ is therefore applied to the composite
loss before back-propagation:
\begin{equation}
\mathcal{L}^k_{\mathrm{total}} =
w_k \bigl(
  \mathcal{L}^k_{\mathrm{deno}}
  + \mathcal{L}^k_{\mathrm{diff}}
  + \mu(e)\,\mathcal{L}^k_{u}
\bigr),
\label{eq:total_loss}
\end{equation}
where $\mu(e)$ is a sigmoid ramp-up function~\cite{wang2024genericssl}
and $w_{\mathrm{LA}}{=}1.5$,
$w_{\mathrm{MMWHS}}{=}w_{\mathrm{MNMS}}{=}1.0$.
The weight $1.5$ is selected by grid search on the validation set over
$w_{\mathrm{LA}}\in\{1.0,1.5,2.0,3.0\}$; values beyond $1.5$ recover
LA accuracy at the cost of MNMS generalisation, reflecting the
fundamental trade-off between task-specific performance and shared
representation quality.
Crucially, re-weighting rescales loss magnitudes before differentiation,
preserving the natural gradient directions that SATC and TTACD are
calibrated to operate on---a property that would be violated by
gradient-projection methods such as PCGrad~\cite{yu2020projecting}.

Together, the four components---11-channel physical decoupling,
SNR-adaptive temporal gating (SATC), task-type-aware token dropout
(TTACD), and gradient-balanced loss re-weighting---constitute an
integrated optimisation strategy that addresses semantic, representational,
and gradient-level conflicts within a single parameter set, without
introducing task-specific subnetworks or architectural branching.

\section{Experiments}
\label{sec:experiments}

\subsection{Datasets and Evaluation Protocol}

Three public cardiac benchmarks cover distinct clinical learning
paradigms.
\textbf{LA}~\cite{xiong2021global}: 100 gadolinium-enhanced MR scans
(80/20 train-test split); the standard 10\%-labeled SSL protocol
(8 labeled, 72 unlabeled) is adopted.
\textbf{MMWHS}~\cite{zhuang2016multi}: 20 labeled MR and 20 unlabeled
CT volumes for MR$\to$CT UDA; targets are ascending aorta (AA), left
atrium cavity (LAC), left ventricle cavity (LVC), and myocardium (MYO).
\textbf{MNMS}~\cite{campello2021multi}: four scanner vendors
(Domains A--D); Domains B, C, D provide 5\% labeled training data
and Domain A is the unseen test target.
Dice (\%) and Jaccard (\%) measure volumetric overlap; 95HD (mm) and
ASD (mm) measure boundary accuracy.

\subsection{Implementation Details}

UniT-Diff is built on DiffVNet~\cite{wang2024genericssl} with an
11-channel output head and trained for 300 epochs on a single
NVIDIA RTX 5090 (24\,GB) using SGD (momentum 0.9,
weight decay $3{\times}10^{-5}$, lr 0.01, poly schedule) with AMP.
A multi-dataloader samples one mini-batch per task per iteration;
dataset-length imbalance is corrected by sample repetition.
Patch sizes are $112{\times}112{\times}80$ (LA),
$128{\times}128{\times}128$ (MMWHS), $32{\times}128{\times}128$ (MNMS).
Hyper-parameters $\tau_k$ and $w_{\mathrm{LA}}{=}1.5$ follow the
ablation in Sect.~\ref{subsec:ttacd}.

\subsection{Comparison with State-of-the-Art}

\subsubsection{Semi-Supervised Learning on LA}

\begin{table}[!htb]
\caption{Results on LA (SSL, 10\% labeled). Best in \textbf{bold}.}
\label{tab:la_results}
\centering\small\setlength{\tabcolsep}{5pt}
\begin{tabular}{lcccc}
\toprule
\textbf{Method} & \textbf{Dice}$\uparrow$ & \textbf{Jaccard}$\uparrow$
  & \textbf{95HD}$\downarrow$ & \textbf{ASD}$\downarrow$ \\
\midrule
MC-Net (2021)~\cite{wu2021mcnet}            & 87.62 & 78.25 & 10.03 & 1.82 \\
SS-Net (2022)~\cite{wu2022ssnet}            & 88.55 & 79.62 &  7.49 & 1.90 \\
Simcvd (2022)~\cite{you2022simcvd}          & 89.03 & 80.34 &  8.34 & 2.59 \\
BCP (2023)~\cite{bai2023bcp}                & 89.62 & 81.31 &  6.81 & 1.76 \\
MLRPL (2024)~\cite{su2024mlrpl}             & 89.86 & 81.68 &  6.91 & 1.85 \\
UGPL (2025)~\cite{ugpl2025}                 & 89.95 & 81.81 &  5.73 & 1.79 \\
GenericSSL (2024)~\cite{wang2024genericssl} & 90.31 & 82.40 &  5.55 & 1.64 \\
UniDeg (2025)~\cite{kumari2025unified}      & 91.07 & 83.67 &  4.96 & 1.65 \\
\midrule
\textbf{UniT-Diff (Ours)} & \textbf{91.18} & \textbf{83.85}
  & \textbf{4.82} & \textbf{1.53} \\
\bottomrule
\end{tabular}
\end{table}

UniT-Diff reaches $91.18\%$ Dice (Table~\ref{tab:la_results}),
surpassing the single-task baseline GenericSSL by $0.87\,\text{pp}$
and the unified competitor UniDeg by $0.11\,\text{pp}$.
The $0.71\,\text{mm}$ reduction in 95HD suggests that cross-task
cardiac priors from MMWHS and MNMS sharpen boundary localisation
beyond single-task capacity.

\subsubsection{Unsupervised Domain Adaptation on MMWHS}

\begin{table}[!htb]
\caption{Results on MMWHS (UDA, MR$\to$CT).}
\label{tab:mmwhs_results}
\centering\small\setlength{\tabcolsep}{2.8pt}
\begin{tabular}{lcccccc}
\toprule
\multirow{2}{*}{\textbf{Method}}
  & \multicolumn{4}{c}{\textbf{Dice (\%)}$\uparrow$}
  & \multirow{2}{*}{\textbf{Avg}$\uparrow$}
  & \multirow{2}{*}{\textbf{ASD}$\downarrow$} \\
\cmidrule(lr){2-5}
& AA & LAC & LVC & MYO & & \\
\midrule
SIFA (2020)~\cite{chen2020sifa}              & 81.3 & 79.5 & 73.8 & 61.6 & 74.1 & 7.0 \\
DSFN (2020)~\cite{zou2020dsfn}               & 84.7 & 76.9 & 79.1 & 62.4 & 75.8 & N/A \\
DSAN (2021)~\cite{lu2021dsan}                & 79.9 & 84.8 & 82.8 & 66.5 & 78.5 & 5.9 \\
LMISA-3D (2022)~\cite{jafari2022lmisa}       & 84.5 & 82.8 & 88.6 & 70.1 & 81.5 & 2.3 \\
Diffuse-UDA (2024)~\cite{yang2024addressing} &  --  &  --  &  --  &  --  & 88.30 & 1.60 \\
ReTri (2026)~\cite{Gao2026retri}             &  --  &  --  & 91.99 & 77.53 & 84.76 & \textbf{0.856} \\
GenericSSL (2024)~\cite{wang2024genericssl}  & \textbf{93.2} & 89.5 & 91.7 & 86.2 & 90.10 & 1.7 \\
UniDeg (2025)~\cite{kumari2025unified}       & 85.4 & \textbf{92.9} & 91.0 & \textbf{95.1} & 91.10 & 1.60 \\
\midrule
\textbf{UniT-Diff (Ours)} & 89.8 & 92.1 & \textbf{92.0} & 93.6
  & \textbf{91.87} & 1.60 \\
\bottomrule
\end{tabular}
\end{table}

UniT-Diff achieves $91.87\%$ average Dice (Table~\ref{tab:mmwhs_results}),
a $1.77\,\text{pp}$ gain over GenericSSL and $0.77\,\text{pp}$ over
UniDeg.
MYO rises from $86.2\%$ to $93.6\%$, consistent with the thin
myocardial wall benefiting from the shared denoising prior across all
three datasets.
AA declines from $93.2\%$ to $89.8\%$: the aortic arch borders the
pulmonary trunk, a region unannotated in LA and MNMS, leaving no
corrective gradient signal in the unified setting
(see Sect.~\ref{sec:discussion}).

\subsubsection{Domain Generalisation on MNMS}

\begin{table}[!htb]
\caption{Results on MNMS (DG, 5\% labeled, Dice \%).}
\label{tab:mnms_results}
\centering\small\setlength{\tabcolsep}{5pt}
\begin{tabular}{lccccc}
\toprule
\textbf{Method} & \textbf{A} & \textbf{B} & \textbf{C} & \textbf{D}
  & \textbf{Avg} \\
\midrule
nnUNet (2021)~\cite{isensee2021nnunet}      & 65.30 & 79.73 & 78.06 & 81.25 & 76.09 \\
SDNet+Aug (2019)~\cite{chartsias2019sdnet}  & 71.21 & 77.31 & 81.40 & 79.95 & 77.47 \\
LDDG (2020)~\cite{li2020lddg}              & 66.22 & 69.49 & 73.40 & 75.66 & 71.29 \\
SAML (2020)~\cite{liu2020saml}             & 67.11 & 76.35 & 77.43 & 78.64 & 74.88 \\
BCP (2023)~\cite{bai2023bcp}               & 73.66 & 79.04 & 77.01 & 78.49 & 77.05 \\
DGNet (2021)~\cite{liu2021dgnet}           & 72.40 & 80.30 & 82.51 & 83.77 & 79.75 \\
TEGDA (2025)~\cite{tegda2025}              &  --   & 83.78 & 79.34 & 82.04 & 81.72 \\
vMFNet (2022)~\cite{liu2022vmfnet}         & 77.06 & 82.29 & 84.01 & \textbf{85.13} & 82.12 \\
GenericSSL (2024)~\cite{wang2024genericssl}& 81.71 & \textbf{85.44} & 82.18 & 83.90 & 83.31 \\
\midrule
\textbf{UniT-Diff (Ours)} & \textbf{82.56} & 85.26 & \textbf{84.16}
  & 84.78 & \textbf{84.19} \\
\bottomrule
\end{tabular}
\end{table}

On the unseen Domain~A scanner (Table~\ref{tab:mnms_results}),
UniT-Diff reaches $84.19\%$ average Dice, exceeding GenericSSL by
$0.88\,\text{pp}$ and vMFNet---a dedicated DG method---by
$2.07\,\text{pp}$.
Gains hold across three of four domains ($+0.85$, $+1.98$,
$+0.88\,\text{pp}$ on A, C, D); Domain~B trails GenericSSL by a
marginal $0.18\,\text{pp}$.

\subsection{Ablation Study}

\begin{table}[!htb]
\caption{Component ablation (Avg Dice \%). Baseline: naive 9-channel joint training.}
\label{tab:ablation}
\centering\small\setlength{\tabcolsep}{5pt}
\begin{tabular}{ccccccc}
\toprule
11-cls & SATC & TTACD & $w_{\mathrm{LA}}$
  & \textbf{LA} & \textbf{MMWHS} & \textbf{MNMS} \\
\midrule
\multicolumn{3}{c}{\emph{Naive 9-cls joint}} & & 83.38 & 87.52 & 76.25 \\
\checkmark & & & & 90.29 & 88.94 & 82.34 \\
\checkmark & \checkmark & & & 90.72 & 91.32 & 83.44 \\
\checkmark & & \checkmark & & 90.84 & 91.34 & 83.50 \\
\checkmark & \checkmark & \checkmark & & 90.93 & 91.67 & 83.86 \\
\checkmark & \checkmark & \checkmark & \checkmark
  & \textbf{91.18} & \textbf{91.87} & \textbf{84.19} \\
\bottomrule
\end{tabular}
\end{table}

The 9-channel baseline collapses LA to $83.38\%$; 11-channel expansion
alone recovers most of this loss by eliminating gradient sign reversal.
SATC and TTACD act on the same task-embedding pathway and yield
comparable single-component gains ($+0.55$/$+0.64\,\text{pp}$ on LA,
$+2.38$/$+2.40\,\text{pp}$ on MMWHS), meaning either can partially
compensate for the absence of the other.
Their combination is synergistic rather than additive: SATC gates
temporally while TTACD enforces structural pathway separation,
two orthogonal effects.
Loss re-weighting ($w_{\mathrm{LA}}{=}1.5$) closes the residual LA
gap without affecting MNMS, identifying gradient imbalance between
the 8-scan LA set and the larger MMWHS as the final bottleneck.

\subsection{Qualitative Analysis}

\begin{figure}[!t]
\centering
\setlength{\tabcolsep}{1pt}
\renewcommand{\arraystretch}{0.6}
\begin{tabular}{ccccc}
{\scriptsize\textbf{Task}}
  & {\scriptsize\textbf{2D GT}}
  & {\scriptsize\textbf{2D Pred}}
  & {\scriptsize\textbf{3D GT}}
  & {\scriptsize\textbf{3D Pred}} \\[1pt]
\rotatebox{90}{\;\scriptsize\textbf{LA}}
  & \includegraphics[width=0.085\textwidth]{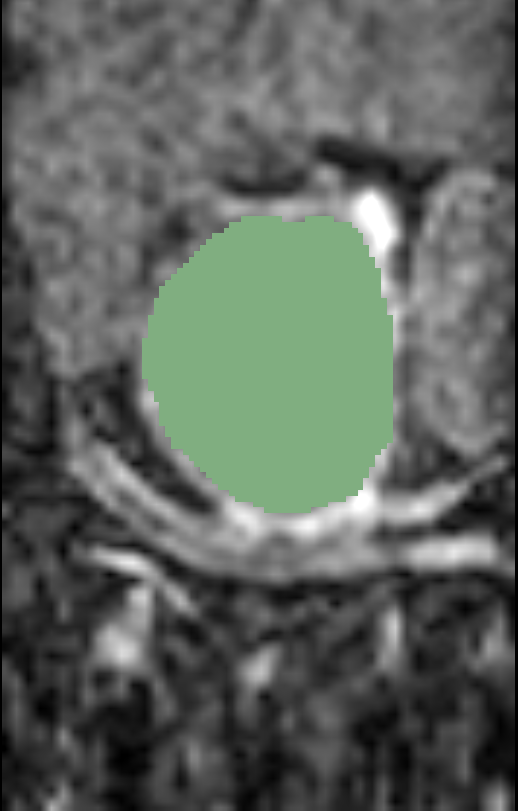}
  & \includegraphics[width=0.085\textwidth]{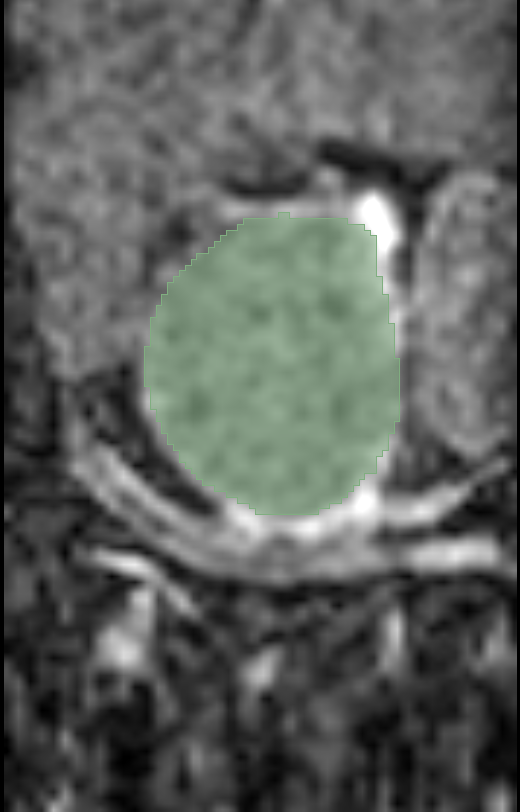}
  & \includegraphics[width=0.085\textwidth]{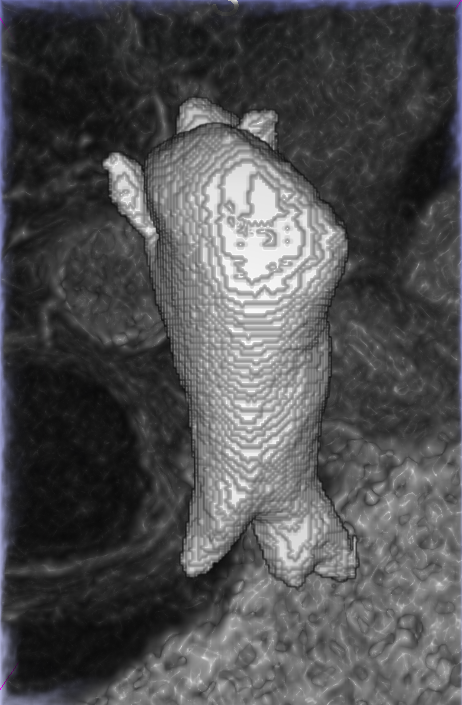}
  & \includegraphics[width=0.085\textwidth]{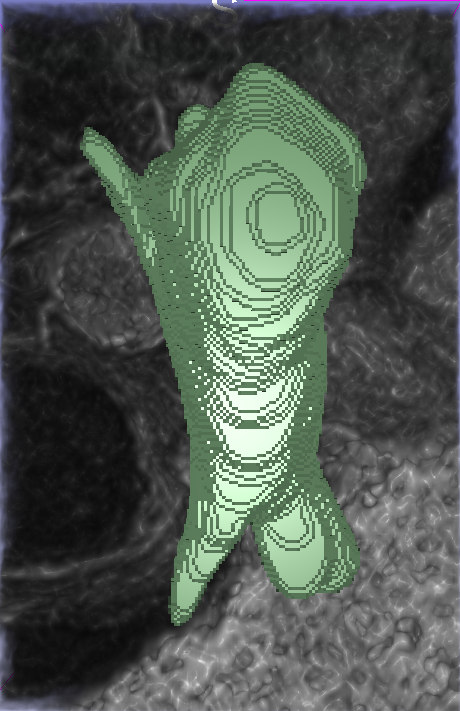} \\[2pt]
\rotatebox{90}{\scriptsize\textbf{MMWHS}}
  & \includegraphics[width=0.085\textwidth]{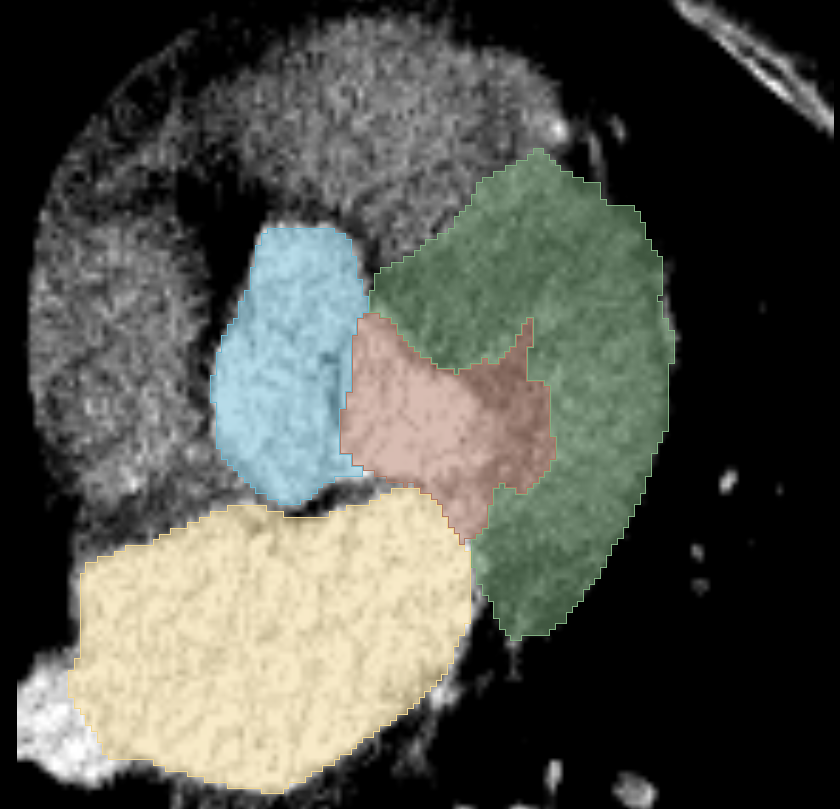}
  & \includegraphics[width=0.085\textwidth]{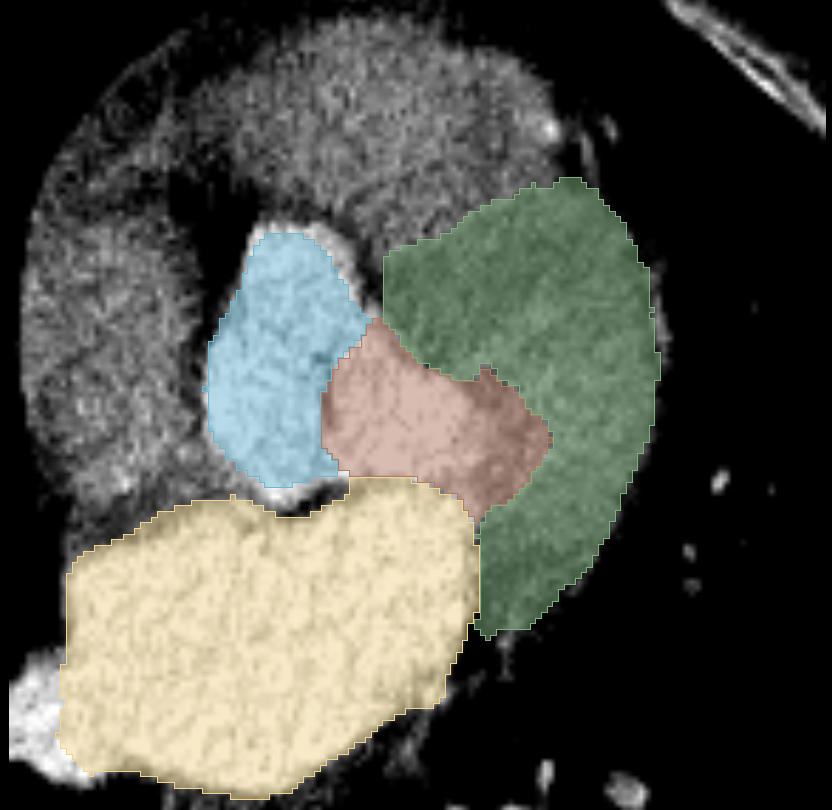}
  & \includegraphics[width=0.085\textwidth]{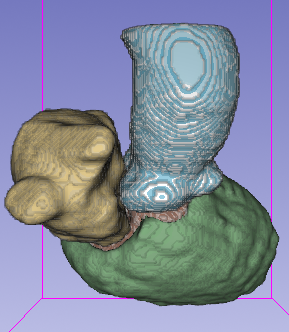}
  & \includegraphics[width=0.085\textwidth]{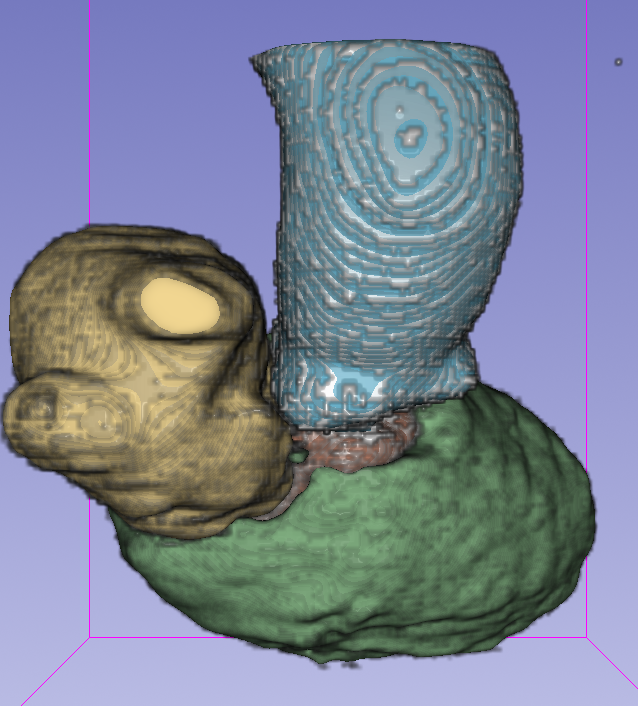} \\[2pt]
\rotatebox{90}{\;\scriptsize\textbf{MNMS}}
  & \includegraphics[width=0.085\textwidth]{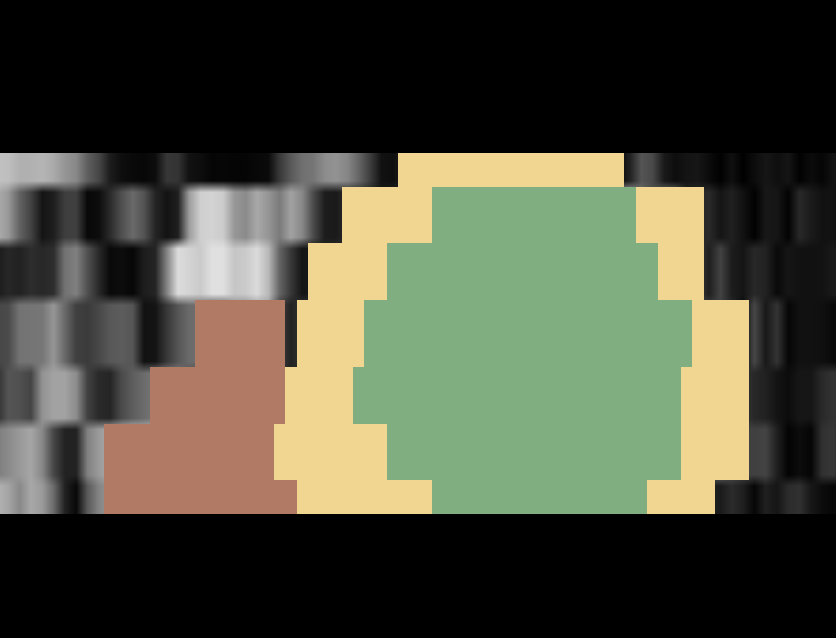}
  & \includegraphics[width=0.085\textwidth]{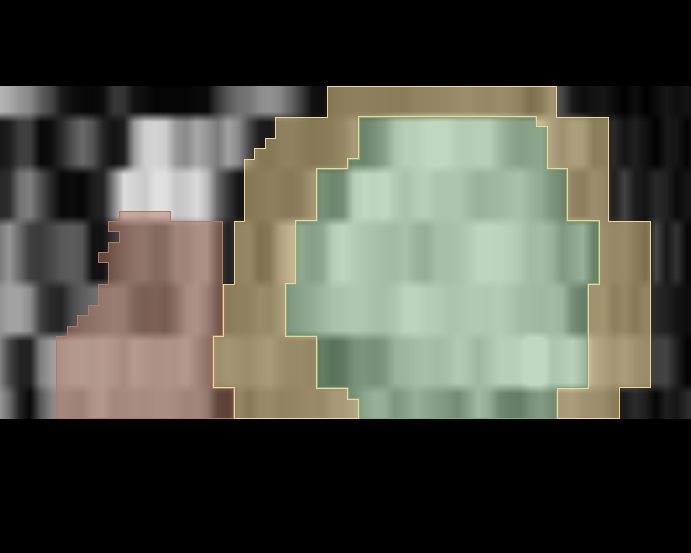}
  & \includegraphics[width=0.085\textwidth]{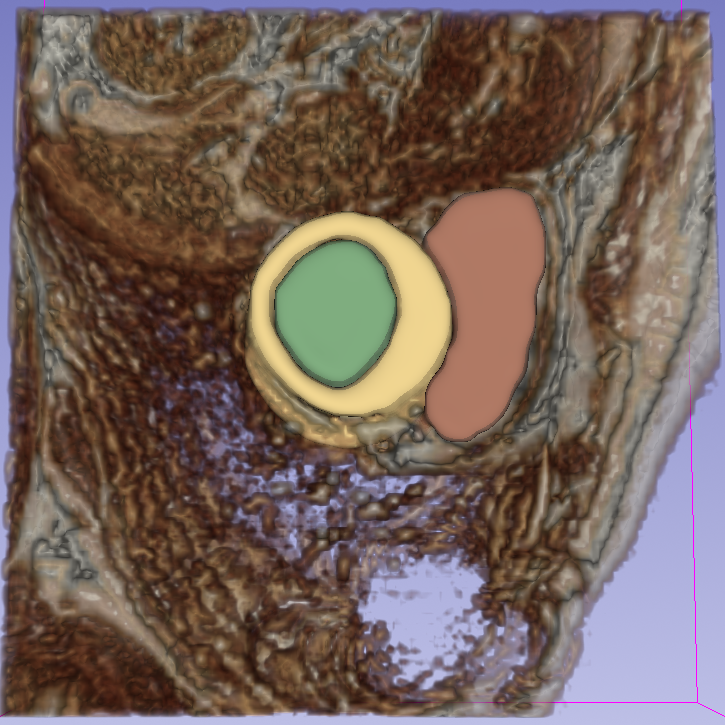}
  & \includegraphics[width=0.085\textwidth]{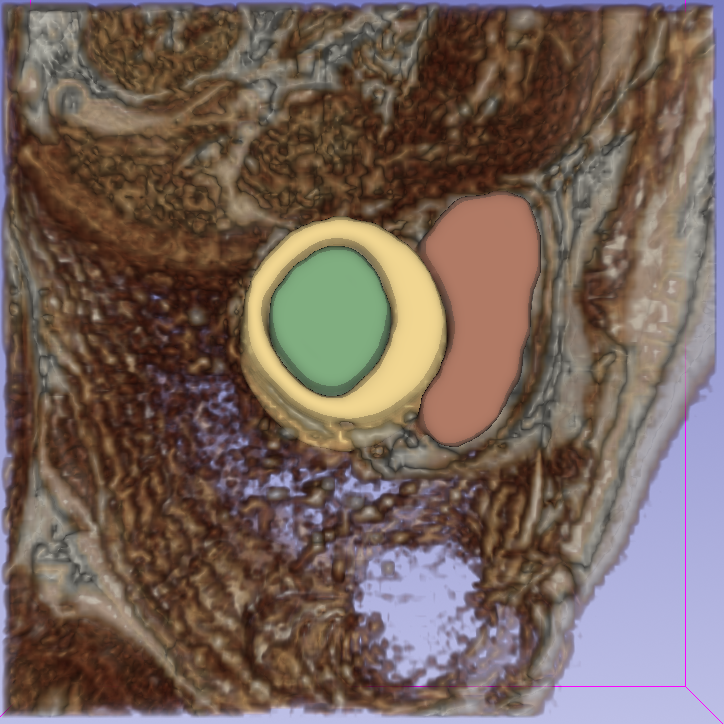} \\
\end{tabular}
\caption{Qualitative results (2D GT / Pred, 3D GT / Pred renders).
\textbf{LA}: no false positives in adjacent ventricular tissue,
confirming semantic collision is eliminated.
\textbf{MMWHS}: all four substructures correctly separated; thin
myocardium recovered with high fidelity ($+7.4\,\text{pp}$ MYO
vs.\ GenericSSL).
\textbf{MNMS}: biventricular shape preserved on unseen Domain~A,
with boundary roughness only in high-gradient slices.}
\label{fig:qualitative}
\end{figure}

Fig.~\ref{fig:qualitative} complements the quantitative results.
On LA, no false positives extend into adjacent ventricular tissue,
confirming 11-channel decoupling suppresses semantic collision.
On MMWHS, the thin myocardial wall is reconstructed with high surface
fidelity despite the MR$\to$CT gap; the ascending aorta is slightly
under-extended at the arch boundary, consistent with the AA regression.
On MNMS, the neutral pathway produces anatomically plausible
reconstructions on Domain~A without test-domain exposure.

\subsection{Discussion}
\label{sec:discussion}

Ablation results follow a consistent ordering: MMWHS gains are large
and stable, MNMS gains are moderate, and LA is most sensitive to
design choices.
MMWHS benefits most because CT/MRI dissimilarity makes task
conditioning directly effective; MNMS profits indirectly through the
cross-dataset cardiac anatomy; LA sits between the two, where
single-modality data limits the benefit of explicit task guidance
while MMWHS gradients subtly reshape the shared encoder.

The AA regression ($93.2\%{\to}89.8\%$ vs.\ GenericSSL) illustrates
an inherent limitation of unified optimisation under annotation
heterogeneity: the aortic arch borders the pulmonary trunk, a region
neither LA nor MNMS annotates, so no corrective gradient exists in
the unified setting.
A second limitation is the false background prior in
$x_{\mathrm{start}}$: inactive channels are padded with $-1$ rather
than masked, constraining denoising capacity on 7--9 of 11 channels.
Dynamic channel routing is a natural extension that would remove this
constraint without restructuring the overall framework.

\section{Conclusion}
\label{sec:conclusion}

UniT-Diff consolidates SSL, UDA, and DG cardiac segmentation into a
single parameter set by resolving task conflicts at three levels:
label space, timestep conditioning, and token dropout policy.
Two findings generalise beyond the evaluated benchmarks.
Task conditioning in diffusion models is not uniformly beneficial:
its value depends on both the noise level and the learning objective,
and withholding the token from DG inputs provides implicit
regularisation through cross-dataset cardiac priors that single-task
training cannot replicate.
The per-task temperature in SATC converges to distinct values across
tasks, offering a diagnostic measure of how much each task benefits
from explicit manifold guidance.
The primary limitation---inactive channels padded with $-1$ rather
than masked---introduces a false background prior that constrains
denoising capacity; dynamic channel routing is the natural remedy.
Extending the framework to broader anatomical domains and larger task
repertoires remains future work.

\end{document}